# Toward Knowledge-Guided AI for Inverse Design in Manufacturing: A Perspective on Domain, Physics, and Human-AI Synergy

*Hugon Lee*[1†], *Hyeonbin Moon*[1†], *Junhyeong Lee*[1†], *and Seunghwa Ryu*[1,2*]

[1]Department of Mechanical Engineering, Korea Advanced Institute of Science and Technology (KAIST), 291 Daehak-ro, Yuseong-gu, Daejeon 34141, Republic of Korea

[2]KAIST InnoCORE PRISM-AI Center, Korea Advanced Institute of Science and Technology (KAIST), Daejeon, 34141, Republic of Korea

**Funding**: National Research Foundation(NRF) of Korea, Grant/Award Number: RS-2023-00222166 and RS-2023-00247245

**Keywords**: inverse design, physics-informed machine learning, large language models, design automation, human-AI collaboration, smart manufacturing

---

[†] Equal contribution.

[*] Corresponding author: ryush@kaist.ac.kr

**Abstract**

Artificial intelligence (AI) is reshaping inverse design in manufacturing, enabling high-performance discovery in materials, products, and processes. However, purely data-driven approaches often struggle in realistic manufacturing settings characterized by sparse data, high-dimensional design spaces, and complex constraints. This perspective proposes an integrated framework built on three complementary pillars: domain knowledge to establish physically meaningful objectives and constraints while removing variables with limited relevance, physics-informed machine learning to enhance generalization under limited or biased data, and large language model-based interfaces to support intuitive, human-centered interaction. Using injection molding as an illustrative example, we demonstrate how these components can operate in practice and conclude by highlighting key challenges for applying such approaches in realistic manufacturing environments.

## 1. Introduction

Artificial intelligence (AI) has expanded the capabilities of inverse design in manufacturing, enabling the creation of design configurations that meet targeted performance requirements with reduced reliance on costly simulations and trial-and-error experimentation. Surrogate modeling techniques—such as neural networks (NNs) and Gaussian processes (GPs)—have been widely adopted to approximate complex physical behaviors. Supported by universal approximation theorems,[1,2] these models can, in principle, capture highly nonlinear relationships between design variables and performance metrics with arbitrary precision, making them attractive tools for modeling and optimization in engineering context.

In recent years, AI applications have extended beyond predictive analytics into generative and prescriptive domains, where the objective is to discover novel material formulations,[3–5] product geometries,[6–9] or process parameters.[10–13] This inverse design paradigm has been empowered by advances in surrogate-assisted optimization, reinforcement learning, and generative modeling.[12,14–16] However, despite the increasing capacity of machine learning (ML) models, their performance in real-world manufacturing environments often face significant challenges, including sparse data availability, high-dimensional design spaces, stringent manufacturing constraints, and the need for interpretability.[10–12] These limitations highlight the need for frameworks that integrate domain expertise and physical reasoning rather than relying solely on unconstrained data-driven approaches.

Addressing these challenges requires more than improving learning algorithms alone. A critical first step is expert-guided problem definition, in which domain knowledge is applied to establish meaningful objectives, define feasible operating ranges, and eliminate design parameters that are irrelevant or weakly coupled to performance outcomes. This process reduces the complexity of the design space, increase the relevance of generated data, and improves the likelihood that AI models will produce solutions that are both optimal in theory and feasible in practice. Physics-informed machine learning (PIML) complements this approach by embedding governing physical principles into the learning process, improving generalization and reliability when data are scarce. Large language models (LLMs) further extend accessibility through natural language (NL) interfaces, enabling more intuitive interaction with design workflows and facilitating communication between human experts and AI systems.

While each of these components—problem definition, PIML, and human-AI interaction—has been studied independently,[15,17–22] their combined application to inverse design in manufacturing remains underexplored. This article suggests that integrating these elements into a cohesive framework can improve data efficiency, generalization, and interpretability.

The remainder of this article is organized as follows. Section 2 briefly reviews AI-driven inverse design process. Section 3 discusses the role of expert-guided problem definition in establishing meaningful objectives and constraints. Section 4 examines PIML methods for embedding physical priors into learning process. Section 5 considers LLM-based interfaces for enabling further generalization and intuitive human interaction. Section 6 illustrates these ideas through the case of injection molding, and Section 7 concludes with a discussion of open challenges for advancing AI-driven inverse design in manufacturing.

## 2. AI-based inverse design

We begin by briefly reviewing the common practices in AI-based inverse design. ML models, also referred to as data-driven models, surrogate models, or metamodels, approximate complex problem by solely utilizing given input-output data pairs (training data). The models are excellent in deriving fast prediction for given input variables and thus can effectively utilized with many inverse design/optimization algorithms. The representative workflow is shown in Figure 1, highlighting where AI methods contribute, from formulating the problem to validating optimized outcomes.

The process begins with problem definition, where design objectives, variables, and constraints are established. This stage is critical as it determines the boundaries of the design space and ensures that subsequent steps focus on feasible and meaningful targets. Once defined, data acquisition supplies the information base for modeling, typically involving experimental measurements and/or simulation outputs. The decision on which design variables to evaluate outputs are typically conducted via design of experiment (DoE) methods.[a]

With data available, ML forward modeling constructs predictive surrogates of the underlying physical processes. Depending on data availability and the need for physical consistency, this may rely on conventional ML, PIML, or emerging LLMs. Inverse design proceeds via two complementary approaches. In the first, an explicit forward model is coupled with optimization methods such as genetic algorithm (GA), Bayesian optimization (BO), or reinforcement learning (RL) to search for optimal designs. In the second, direct inverse design bypasses explicit forward modeling by using generative models[b] or LLMs to map desired properties directly to candidate designs. The resulting optimized design may undergo a validation stage, where predictions are tested against experiments or high-fidelity simulations to ensure reliability before practical deployment. More comprehensive treatment on AI-based inverse design can be found in recent review papers.[15,22–24]

[a] Representative DoE methods include grid sampling, random sampling, Sobol sampling, and Latin hypercube sampling.

Within this framework, each component discussed throughout this *Perspective* plays a different role: expert knowledge shapes problem definition (Section 3), PIML improves forward modeling under sparse or biased data (Section 4), and LLMs enhance usability and interaction, acting both as predictors/generators and interfaces (Section 5). These concepts are further demonstrated in an injection molding case study (Section 6).

**3. Importance of expert-guided problem definition**

Inverse design in manufacturing inherently involves navigating complex, high-dimensional design spaces under strict performance requirements and complex process constraints.[7,10–12,16] A critical challenge in this context is that both experimental and simulation-generated datasets are often limited or biased in terms of operating conditions, material classes, or part geometries.[24,25] Such bias can arise because experimental data are typically collected within established process windows due to practical and economic constraints, while simulation data often rely on simplified physics or restricted parameter ranges to reduce computational cost. As a result, training data may cover only a narrow portion of the feasible design space, leading models to fit well locally but fail to generalize beyond it, thereby limiting their ability to propose optimal or innovative solutions.

---

[b] Representative generative models include variational autoencoders, generative adversarial networks, and diffusion models.

Figure 2 illustrates this issue with simplified two-variable design space with single objective function, where the true function is visualized in Figure 2a.[c] Even with the same number of training samples, different sampling patterns can lead to markedly different predictions even through the ML models fit the available data closely (Figure 2b). While the grid sampling broadly covers the design space and captures global optimum (Figure 2c), random (Figure 2d) or localized (Figure 2e) sampling may lead to become trapped in local optimum point, overlooking better solutions elsewhere.

From this perspective, the interplay between the size of the design space and the coverage of training data is a decisive factor in determining ML model performance, as discussed in recent review.[15] As shown in Figure 3, inverse design scenarios can be broadly classified into four categories. In interpolation problems (Figure 3a), dense and uniform data coverage ensures model credibility and allows direct application of standard global optimization techniques. Extrapolation problems (Figure 3b) refer to cases where training data occupy a small subregion of a feasible space, due to large design space and/or limited data. Small-data problems (Figure 3c) are when the number of available data is so small that the optimum search cannot rely solely on poor ML model predictions. Finally, multi-fidelity problems (Figure 3d) are equipped with low-fidelity and high-fidelity data sources. This *Perspective* focuses on extrapolation setting, which is prevalent in manufacturing applications. Further details on methods for each cases can be found in Lee *et al.*[15]

In the extrapolation scenarios with large design space and/or limited data, expert-guided problem definition can play a critical role. By applying domain knowledge, objectives may be more clearly articulated and design variables reduced to those that are more physically relevant to the target performance. Such reduction can improve data efficiency and alleviate dimensionality issues, but it should ideally be grounded in physical reasoning—for example, established cause-effect relationships, sensitivity analyses, or governing equations. Otherwise, aggressive reduction carries the risk of excluding variables that, while seemingly minor in the available dataset, could become influential under different operating conditions, potentially compromising model robustness.

---

[c] The illustrative two-input one-output problem is the shifted-rotated Rastrigin function,[26] a common benchmark for evaluating ML model performance. The surrogate model used is GP regression with radial basis function kernel.[27]

Complementing variable reduction, the explicit incorporation of manufacturing constraints is equally important. Constraints that can be expressed in limits or functional forms—such as discrete machine settings, processing ranges dictated by material behavior, or geometric limitations of molds or tooling—should be embedded into the problem definition from the outset. Addressing these considerations early helps avoid infeasible design exploration and ensures practical relevance. For instance, in polymer additive manufacturing, improper process conditions can lead to build failure or weak interlayer bonding, highlighting the need to establish feasible variable ranges during problem formultion.[28] Further examples can be found in Section 6.

In summary, expert-guided problem definition—when applied with careful attention to the physical underpinnings of the system—provide a disciplined way to transform an unconstrained, high-dimensional design challenge into a tractable and relevant problem. By clarifying objectives, pruning variables with physically justified reasoning, and embedding manufacturing constraints, this approach enables subsequent data acquisition and modeling to operate more effectively and reliably, while safeguarding against oversimplification that could undermine design quality.

## 4. PIML for data efficiency and reliable generalization

By applying expert-guided problem definition, the design space is reshaped into a physically meaningful and practically feasible domain, with unnecessary variables removed and constraints embedded from the outset. However, even with this careful formulation, manufacturing datasets often remain sparse, noisy, and biased toward limited operating regions. Under such conditions, purely data-driven models may still struggle to generalize or maintain physical consistency. This motivated the integration of PIML, which combines domain-governing equations with statistical learning to improve model robustness, interpretability, and reliability in data-scarce regimes.[14,20,21,29]

PIML combines the strengths of scientific computing and data-driven modeling, improving data efficiency and enhancing model generalization while maintaining interpretability (Figure 4a). By constructing predictions to obey known physics, PIML can operate effectively in data-scarce regimes, while still leveraging data to capture unmodeled effects. This approach has been widely applied to forward and inverse problems, reconstruction of hidden states from sparse observations,[30,31] and acceleration of computationally intensive simulations.[32] Two representative frameworks are physics-informed neural networks (PINNs) and operator learning architectures, which differ in formulation and scalability.

PINNs embed governing partial differential equation (PDE) residuals into the loss function of a neural network (Figure 4b).[33] Inputs typically consist of spatial ($x$ and $y$) and/or temporal ($t$) coordinates, and outputs are physical quantities of interest ($f$) such as displacement or temperature. The loss function combines data loss ($\mathcal{L}_{\text{data}}$) with a physics-based loss term ($\mathcal{L}_{\text{phys}}$) computed via automatic differentiation to enforce PDE constraints.[34] This enables training even in semi-supervised or unsupervised regimes. However, PINNs can be difficult to train for stiff PDEs or high-dimensional problems due to vanishing gradients, loss imbalance, and hyperparameter sensitivity.[35] In inverse design tasks requiring repeated evaluations under varying conditions, retraining is often necessary, reducing their efficiency relative to classical solvers.

Operator learning offers a more scalable alternative by approximating mappings between function spaces rather than fixed-dimensional vectors.[36,37] Operator learning architectures such as deep operator network (DeepONet) consists of a branch network, which processes functional inputs $\mathbf{u}$ (*e.g.*, boundary conditions, material properties), and a trunk network, which processes spatial or temporal coordinates $\mathbf{y}$ (Figure 4c).[36,38] Their outputs are combined to produce the final solution $G_\theta(\mathbf{u})(\mathbf{y})$ where $G$ is the solution operator and $\theta$ is model hyperparameters. This formulation allows generalization across boundary conditions and geometries without retraining. Physics-informed variants, such as physics-informed DeepONet (PIDON), incorporate PDE residuals in a manner analogous to PINNs.[38,39] While operator models in purely data-driven format typically train more stably than PINNs, they require large, diverse datasets and may generalize poorly to out-of-distribution cases. Other notable operator learning methods include Fourier neural operator (FNO)[40] and physics-informed neural operator (PINO).[41]

In manufacturing settings, PIML has shown promise for data-efficient modeling across complex, physics-driven processes. For example, PINNs have been deployed as virtual thermal sensors in composite manufacturing, reconstructing full-field temperature profiles from sparse in-situ measurements to enable real-time monitoring and defect mitigation.[42] Similar approaches have been applied to polymer processing, where PINNs inferred thermal behavior under incomplete material data.[43] Operator learning models such as DeepONet have proven effective in additive manufacturing of stochastic metamaterials, rapidly evaluating diverse geometries and boundary conditions without retraining,[44] and in metal additive manufacturing, where operator models integrated with sensor feedback adaptively controlled process parameters to improve build quality.[45] Collectively, these cases highlight PIML's ability to maintain physical consistency and generalize under sparse or incomplete data conditions, particularly when applied to problems with well-defined physics and manageable variability.

Despite these advances, significant limitations remain. Many manufacturing problems involve governing equations that are partially known, empirically derived, or too complex to embed directly in learning architectures. Furthermore, PIML methods such as PINNs and operator networks are best suited to continuous, differentiable systems, making them less effective for processes with discrete or categorical inputs common in industry.[46] Real-world manufacturing also features multiphysics coupling, irregular boundaries, and high-dimensional parameters, often leading to numerical stiffness and unstable convergence.[47] Hybrid modeling—which combines partial physics with data-driven learning—offers a promising pathway,[48,49] yet situations frequently arise where physical models are incomplete, or constraints are highly context-specific. In such cases, expert judgement and human-in-the-loop strategies, increasingly supported by LLM-based interfaces, are expected to be promising to guide problem formulation and ensure practical relevance.

## 5. Human-centered design via LLMs

Expert-guided problem definition and physics-informed modeling provide the technical foundation for AI-driven inverse design. Yet, for deployment in manufacturing, enhanced generalization, contextual grounding, and interaction with human remain critical. LLMs are increasingly explored in this role, and their contributions can be organized, but not limited, into two main paradigms: (i) LLMs as predictors or generators of designs and (ii) LLMs as NL interfaces that mediate between human users, knowledge base, and computational tools (see Figure 1).

The LLM applications in domain-specific context can be broadly categorized into three categories (Figure 5): (i) zero- and few-shot prompting (Figure 5a), where pre-trained models are guided by task-specific instructions. (ii) The fine-tuning (Figure 5b) and retrieval-augmented generation (RAG) (Figure 5c), which enhance factual grounding either by adapting the model to domain-specific data or by dynamically retrieving relevant information during inference. (iii) The tool-enhanced and multi-agent architectures (Figure 5d), where LLMs coordinate external tools and/or specialized agents to perform multi-step tasks within a unified workflow.

*LLMs as predictors or generators*

In the first paradigm, LLM itself is utilized as a predictive or generative model within the inverse design process, enabling it to directly generate or evaluate candidate solutions in accordance with contextual and operational constraints. Active lines of research have explored the use of LLMs as structural generation models. For example, CAD-Llama and CAD-Coder were developed through supervised fine-tuning on parametric 3D modeling tasks, enabling the generation of editable CAD code from text or sketches with high accuracy.[50,51] In addition, other approaches have leveraged VLMs to translate sketches into 3D prototypes, thereby expanding design flexibility.[52] Furthermore, FEM-in-the-loop structural design refinement has also been demonstrated, in which candidate structures are generated by LLMs, evaluated using FEM, and iteratively refined to improve performance.[53] The conditional generation capability of LLMs supports adaptation to contextual requirements, and feedback derived from manufacturing constraints can be incorporated by the LLM-based design agent to iteratively refine and propose improved designs. Despite these advances, current methods face several limitations, including limited numerical precision, difficulty in handling complex geometries, and high sensitivity to prompt design.

*LLMs as NL interfaces for inverse design system*

The second approach leverages the LLM as a NL interface, functioning as a form of human-computer interaction (HCI) system.[54–58] HCI focused on aligning users, machines, and services to achieve specific goals, with functionality and usability as central requirements. In this role, LLMs can interpret practitioner requests and either retrieve factual information or invoke domain-specific tools—such as specialized ML models or physics-based simulations—to overcome their limited numerical precision. This strategy enables the integration of existing problem-specific models. Moreover, when requirements extend beyond predefined domain-specific evaluators, the model can still draw upon its pre-trained knowledge to provide plausible responses.

Within this paradigm, RAG strengthens the reliability of LLM-assisted decision making-by dynamically appending relevant documents to user queries—such as datasheets or operating procedures—to user queries, thereby improving accuracy without retraining.[59–62] A representative example is AMGPT, which integrate a Llama2-7B model with a curated additive manufacturing corpus to deliver expert-level responses.[62] Beyond retrieval, NL interfaces can also connect to external tools, enabling LLMs to orchestrate resources such as sensors, controllers, or design software. In this agentic setting, multi-agent LLM architectures distribute specialized roles across agents, supporting workflows such as design optimization, process control, and real-time decision-making.[63–69]

*Limitations and outlook*

Despite these developments, current LLM-based systems for manufacturing still face both technical and practical limitations. On the technical side, many implementations rely on handcrafted prompt templates and heuristic coordination strategies. Although research into automated prompt optimization[70,71] and adaptive agent orchestration[72–74] is emerging, these methods remain in their early stages and are not yet widely adopted in manufacturing contexts. Furthermore, hallucination remains a persistent risk, necessitating rigorous validation and safeguards when applying LLM outputs in high-stakes manufacturing contexts. From a practical standpoint, causality, explainability and accountability are critical in high-consequence applications, necessitating transparent reasoning traces and clear attribution of decisions.[75,76] Moreover, the use of external APIs for fine-tuning or inference on proprietary process data raise concerns regarding intellectual property protection and data privacy. Addressing these issues will require systematic methods for integrating LLMs with physics-based models, advances in automated prompt and agent coordination, and the establishment of standards for security, trust, and transparency.[77,78]

## 6. Illustrative example: injection molding

To illustrate how knowledge-guided AI can be applied in practice, we consider injection molding, one of the most widely used polymer manufacturing processes due to its scalability and ability to produce complex geometries.[79] A simplified process schematics is shown in Figure 6a, where the process is comprised of four stages: filling (injecting molten material into the mold by screw motion), packing (applying pressure ensuring complete filling and compensate thermal shrinkage), cooling (cool the part until solidifies), and extraction (extracting part from the mold). Each stage is influenced by tightly coupled thermal, phase change, mechanical, and flow phenomena, which makes process optimization highly challenging. Small deviations in pressure, temperature, or timing can cause defects[d], dimensional inaccuracies, or degraded mechanical performance, underscoring the importance of systematic design.

[d] Common defects in injection-molded polymer products include short shots (incomplete filling), sink marks (surface depressions), burn marks (caused by overhitting), weld lines (where resin flow fronts meet), burrs (overflowed material), and warpage (part distortion).

Figure 6b highlights three sources of complexity that arise in AI-driven inverse design. First, the number of input parameters is large, including time-dependent injection and packing pressure profiles, resin temperature, and cooling conditions (Figure 6b(i)). Second, numerous manufacturing constraints restrict feasible designs, such as machine operating limits, energy use, allowable material ranges, and limitations on part geometry (Figure 6b(ii), expanded in Figure 6c).[e] Third, environmental variables such as ambient temperature or humidity can shift process behavior and alter optimal parameter settings (Figure 6b(iii)).[f] Table 1 summarizes typical process variables and constraints related to the injection molding process.

Constraints in injection molding extend across multiple layers (Figure 6c): machine limitations (*e.g.*, maximum injection pressure), energy and cost considerations, material-specific requirements such as plasticization and cooling windows, and geometric restrictions in part design (minimum thickness for viscous resin/filler system filling, mold compatibility). These constraints define the feasible design space and should be incorporated into problem formulation.

The inverse design workflow, shown schematically in Figure 6d, demonstrates how the three knowledge-guided components can be integrated. In the first step (Figure 6d(i)), expert-guided problem definition sets proper design variables by excluding variables with limited impact,[12,80] admissible parameter ranges,[10–12] and part design parametrization satisfying geometric constraints,[7] ensuring that optimization focuses on meaningful and feasible designs. In the second step (Figure 6d(ii)), forward ML models are constructed. Here, PIML methods enhance predictive fidelity under sparse or biased data. For instance, PINNs have been applied to weld line prediction by embedding heat transfer and polymer flow equations into the loss function,[81] and for resin viscosity estimation by incorporating rheological constitutive models.[82] These studies illustrate how embedding physics can improve data efficiency and generalization. However, as discussed in Section 4, current PIML methods still face difficulties when handling high-dimensional, multiphysics, and constraint-rich nature of injection molding.

---

[e] The listed constraints are not exhaustive; additional limitations may arise depending on specific material, geometry, or processing conditions.

[f] Environmental variables (*e.g.*, ambient temperature, humidity) affect process outcomes but are typically uncontrollable. They are thus considered as inputs rather than design variables, sometimes referred to as contextual variables.

In the third step (Figure 6d(iii)), the trained models are coupled with optimization or generative approaches to identify design solutions, typically under specified environmental conditions (Figure 6d(iv)). The obtained designs are then validated experimentally or via high-fidelity simulation (Figure 6d(v)) to ensure credibility. To improve adaptability under varying conditions, recent work has begun integrating LLM-based agents. For example, IM-Chat combined RAG with multi-agent coordination to support interactive process design and knowledge transfer, linking documented resources and data-driven process condition generators (Figure 6d).[83] This illustrates the potential of LLMs to act as human-AI interfaces, complementing expert input and physics-based models in iterative design cycles.

Taken together, the case study highlights how expert-guided problem definition, PIML, and LLM-based interfaces offer complementary roles in tackling the complexity of injection molding inverse design. Each pillar addresses distinct challenges—problem formulation, generalization, and human interaction—while their integration provides a pathway toward more reliable and practical design support systems.

## 7. Conclusion and future perspective

AI-driven inverse design in manufacturing is evolving from a purely algorithmic exercise into a knowledge-guided process that integrates domain expertise, physics-based modeling, and human-AI interaction. In this *Perspective*, we identified three complementary pillars that support this shift: expert-guided problem definition, PIML, and LLM-based NL interfaces. The injection molding case study illustrated how these elements can be combined to address the challenges of biased and sparse datasets, high-dimensional parameter spaces, and stringent manufacturing constraints.

Looking ahead, progress in this field will require developing cohesive design ecosystems in which these components operate in concert rather than as isolated tools. Expert knowledge can play a central role in defining physically grounded variables and embedding feasibility constraints, while PIML offers data-efficient and physics-consistent predictions even under sparse or biased data. At the same time, LLM-based interfaces hold potential for interpretable, collaborative interaction with design workflows, provided that reliability and accountability are maintained. Advancing this vision will likely require improved frameworks for variable reduction and constraint embedding, hybrid modeling strategies that integrate partial physics and empirical knowledge, and systematic approaches for benchmarking and governance under realistic manufacturing conditions.

By framing inverse design as a collaborative process between humans and AI, the role of AI shifts from that of a black-box optimizer to a partner in co-creation. Achieving this vision will require not only methodological advances but also careful consideration of interface design, validation practices, and deployment strategies suited to industrial contexts. If these directions are pursued, integrated frameworks can enhance the efficiency, robustness, and applicability of inverse design, thereby expanding opportunities for innovation across diverse manufacturing domains.

## Acknowledgements

This work was supported by the National Research Foundation of Korea(NRF) grand funded by the Korea government(MSIT) (No. RS-2023-00222166 and No. RS-2023-00247245) and the InnoCORE program of the Ministry of Science and ICT(N10250154)..

## Conflicts of interest

The authors declare no conflict of interest.

## Data availability statement

No new data were created or analyzed in this study.

**Figures and tables**

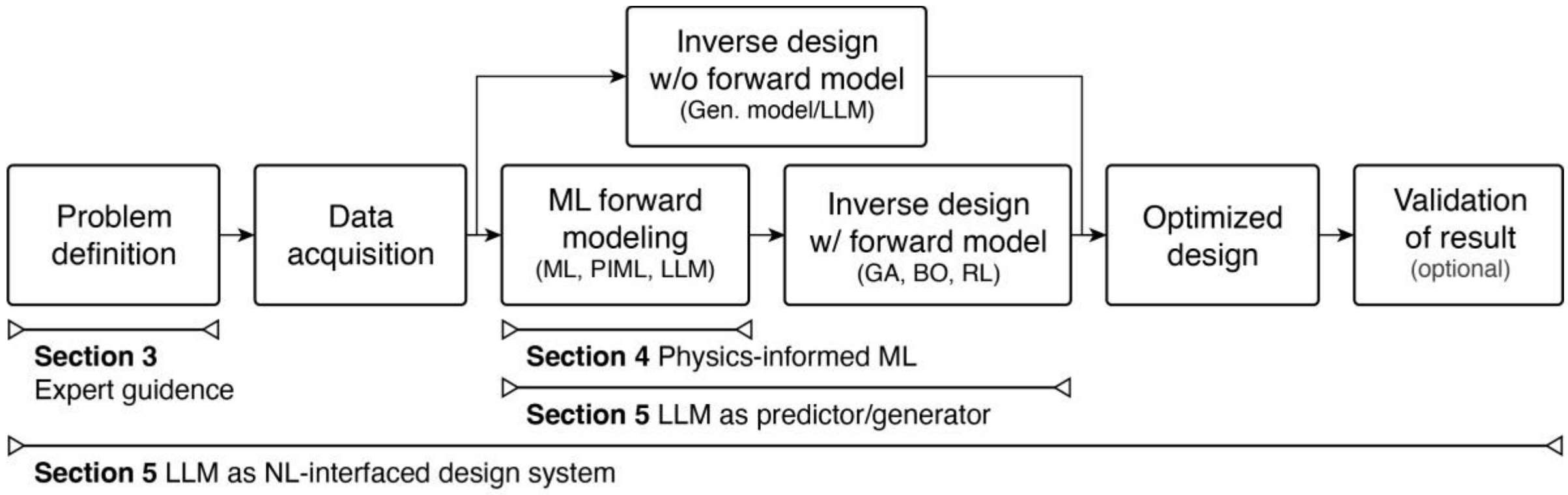

**Figure 1.** Schematic workflow of AI-based inverse design process, highlighting the stages where AI contributes. Concepts discussed in this *Perspective*—expert-guided problem definition (Section 3), PIML (Section 4), and LLM-based interfaces (Section 5)—are marked within the workflow.

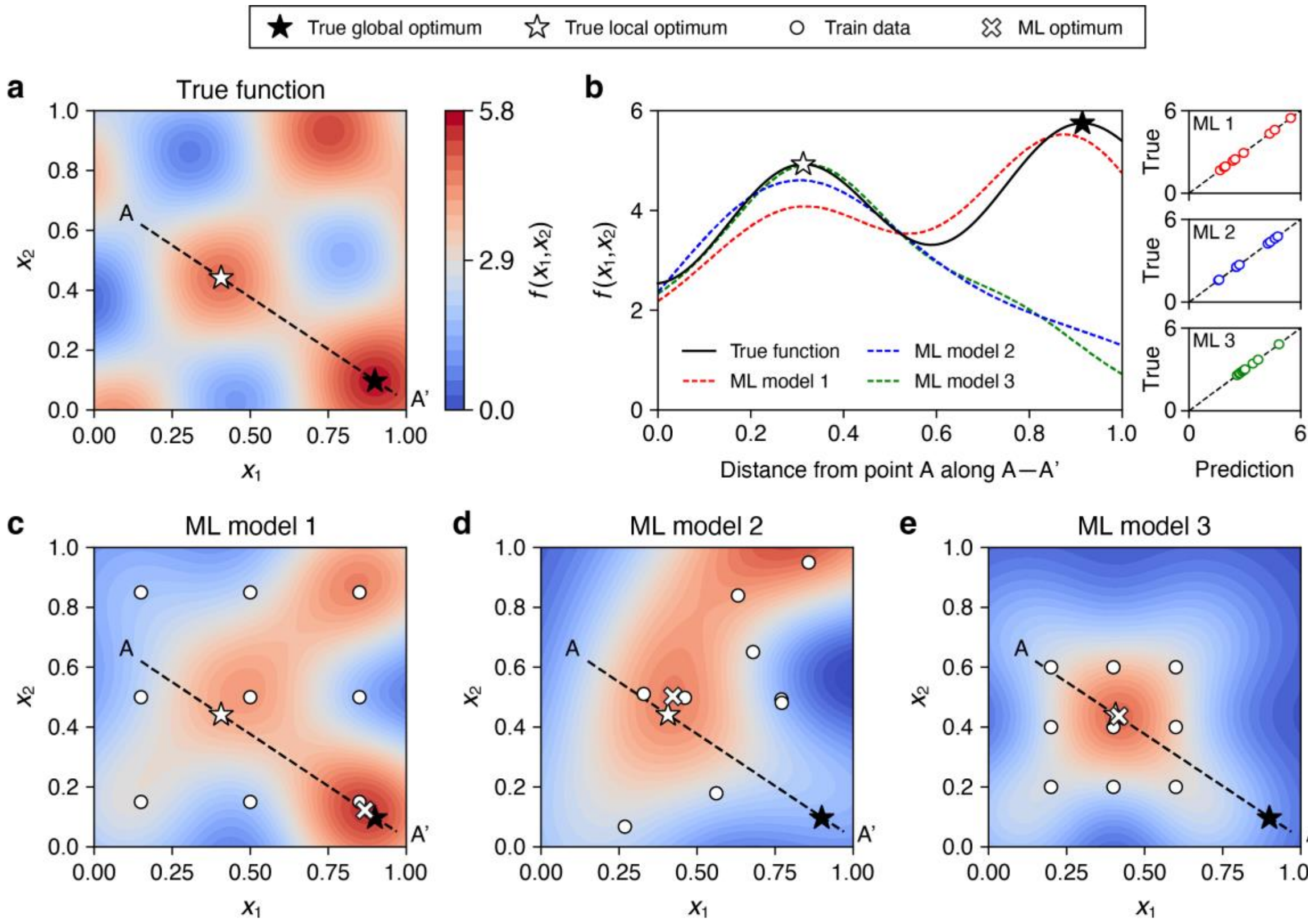

**Figure 2.** Illustration of how training data distribution affects surrogate model predictions in a two-input, single-output problem. The line A—A' passes through the optimal points. (a) Ground-truth function with global and local optima. (b) Comparison of ground truth and ML predictions along line A—A', showing how models can fit training data (panels of right-hand side) but diverge in unseen regions. (c) Grid sampling covers the design space broadly and recovers the global optimum. (d) Random sampling yields uneven coverage and risks missing the global optimum. (e) Localized sampling around a subregion captures only the local optimum.

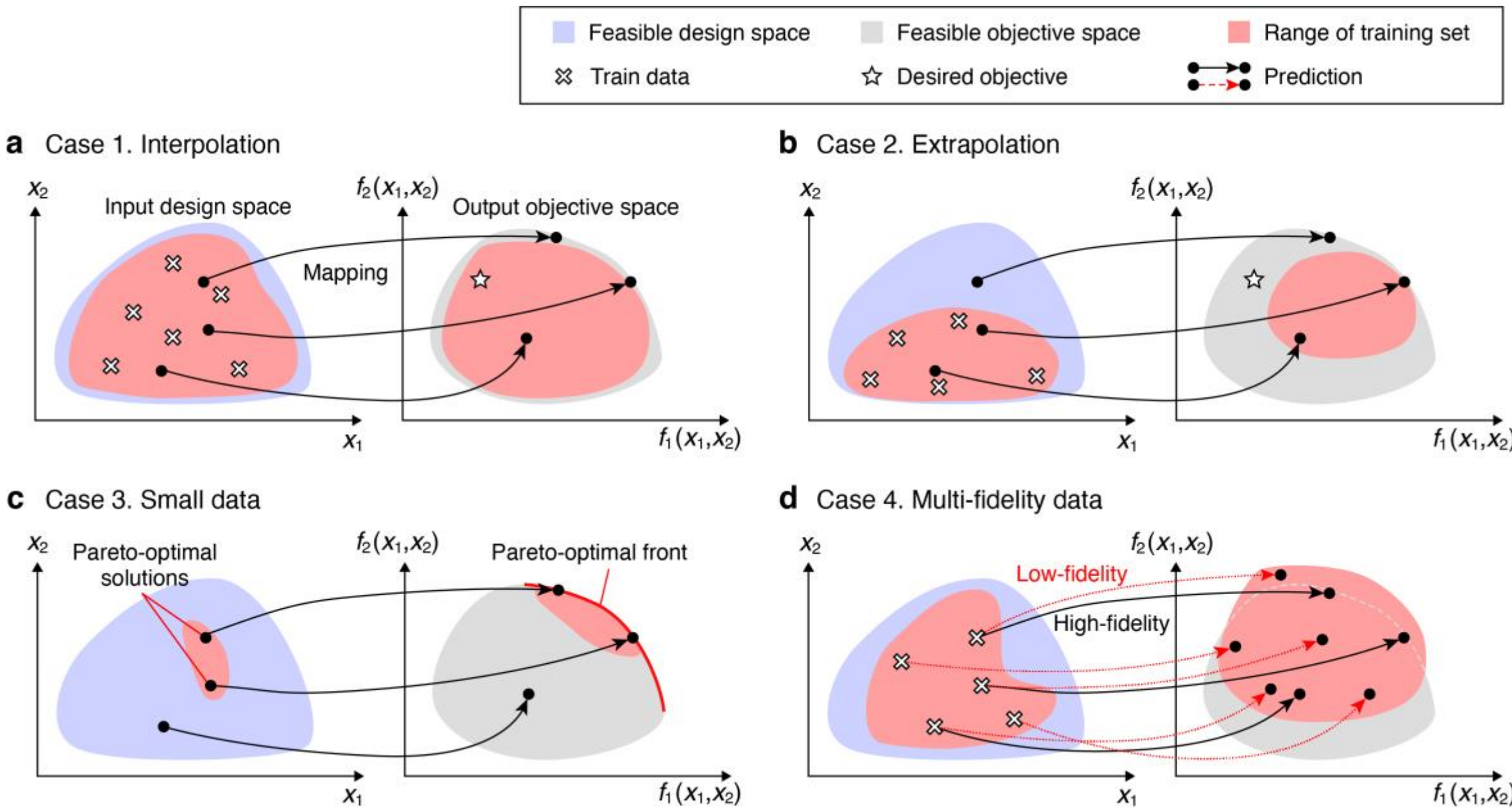

**Figure 3.** Four categories of ML scenarios in engineering inverse design. (a) Interpolation: dense and uniform data coverage enables reliable surrogate modeling and optimization. (b) Extrapolation: training data covers only a small subregion of a large feasible space, requiring strong priors for generalization. (c) Small-data: extremely limited, high-cost datasets necessitate strategic use of every data point. (d) Multi-fidelity data: combining low-fidelity and high-fidelity data sources to balance cost and accuracy.

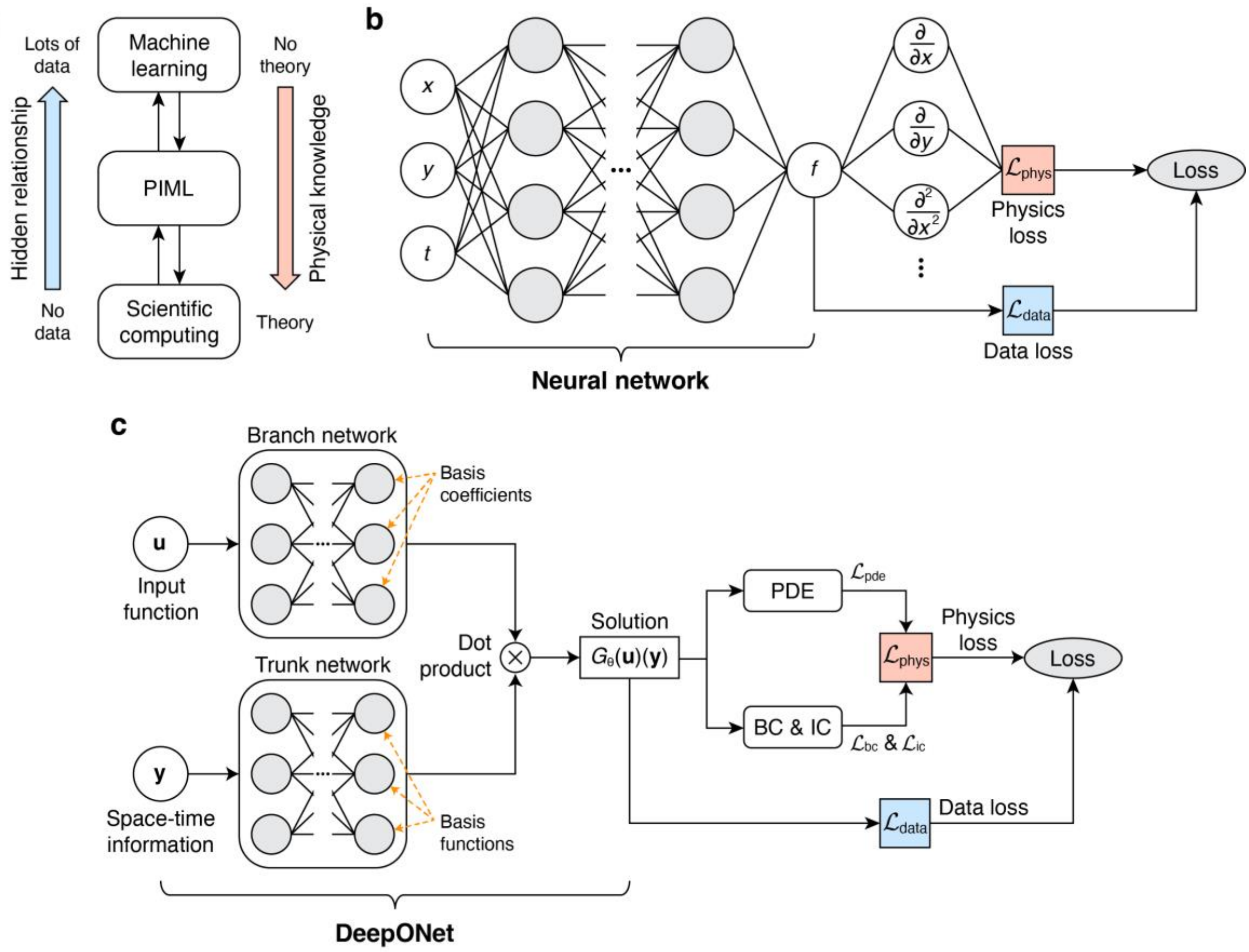

**Figure 4.** Representative PIML models. (a) General PIML framework integrating physics and data to enhance model robustness and generalization. (b) PINNs, which embed PDE residuals into the loss function to enforce governing equations. (c) PIDON, which augments DeepONet with physics-based loss to learn solution operators across varying boundary conditions and geometries.

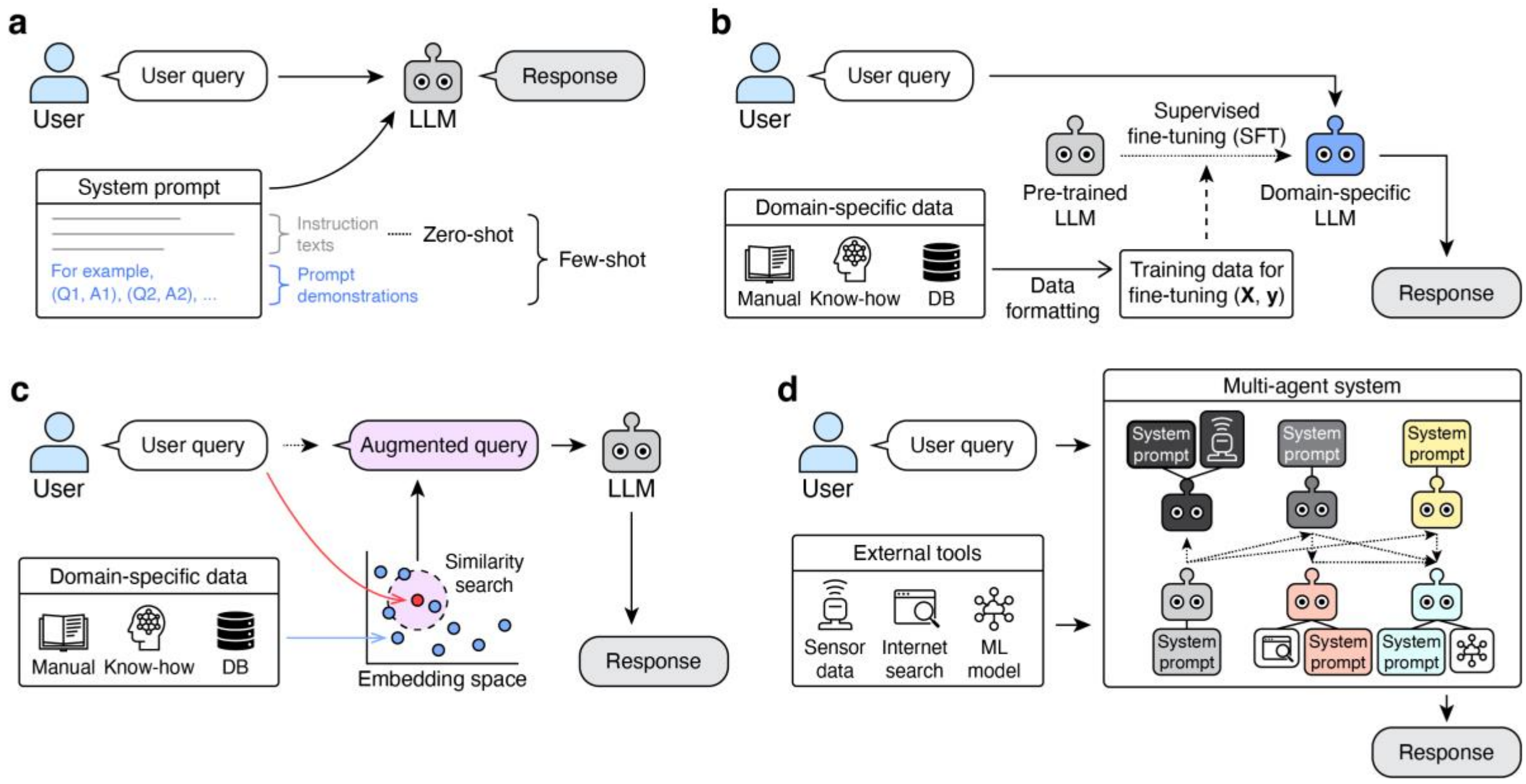

**Figure 5.** Techniques for leveraging LLMs in domain-specific tasks. (a) Zero- and few-shot prompting: guiding pre-trained models with task-specific instructions or Q&A demonstrations. (b) Supervised fine tuning (SFT): adapting pre-trained LLMs using labeled domain-specific datasets for more accurate responses. (c) RAG: dynamically retrieving domain-relevant knowledge (e.g., manuals, know-how, databases) and appending it to queries for improved grounding. (d) Multi-agent systems: coordinating multiple specialized agents and external tools (e.g., sensors, web search, ML models, design software) to perform complex workflows in a modular, collaborative manner.

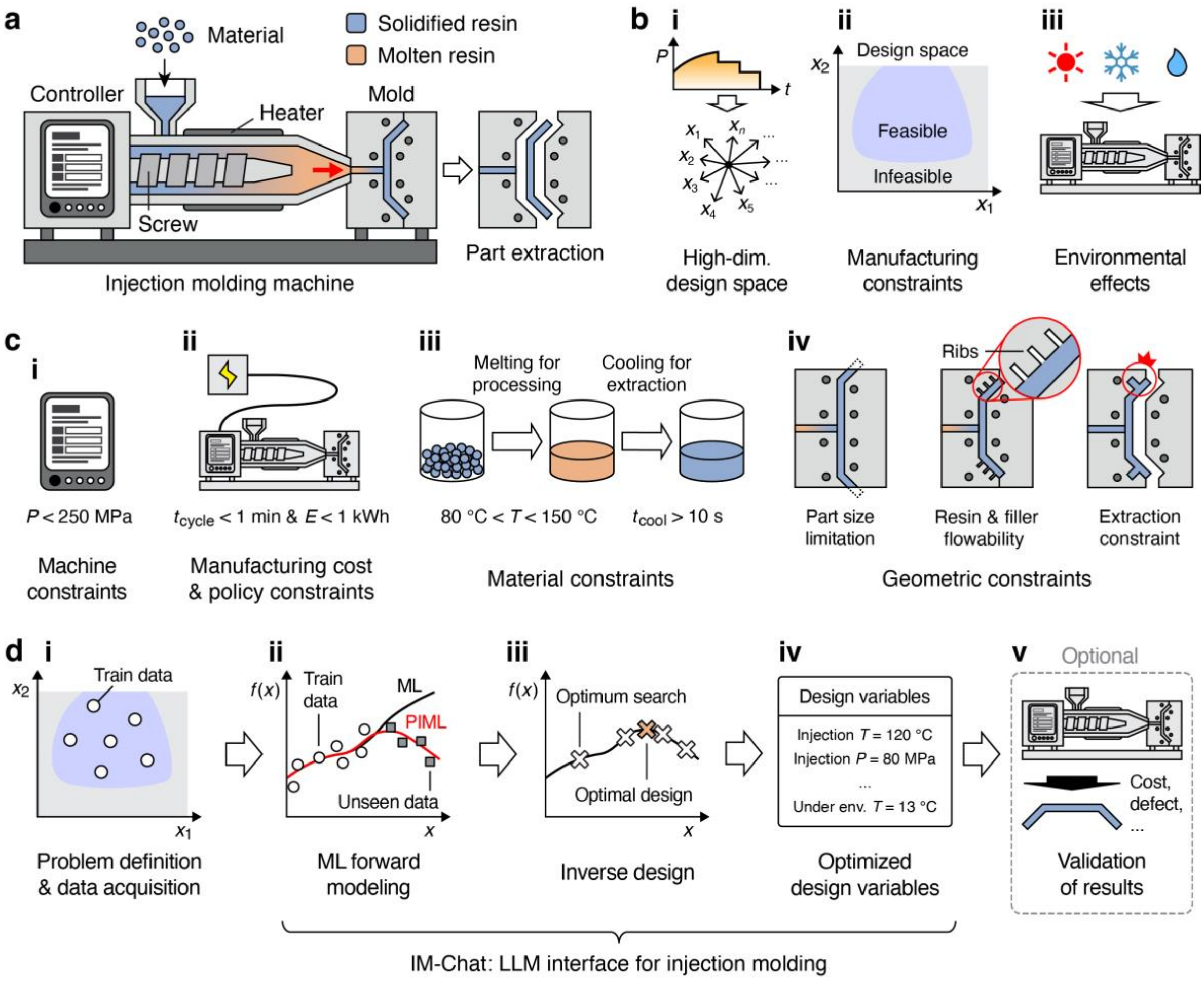

**Figure 6.** AI-based inverse design of injection molding. (a) Process schematics of injection molding. (b) Challenges in AI-driven inverse design: (i) high-dimensional input parametrization, (ii) manufacturing constraints limiting feasible solutions, and (iii) environmental conditions affecting outcomes. (c) Representative manufacturing constraints: (i) machine operating ranges, (ii) energy cost and policy limits, (iii) resin thermal properties and cooling requirements, and (iv) geometric constraints on part design. (d) Inverse design workflow: (i) expert-guided problem definition and DoE-based data acquisition, (ii) training forward ML models with PIML for improved fidelity, (iii) optimization for inverse design, (iv) optimized design variables under specific environmental conditions, and (v) validation of optimized designs. Emerging applications such as LLM-based multi-agent frameworks (e.g., IM-Chat) enable interactive design and knowledge transfer.

**Table 1.** Representative variables and constraints in injection molding inverse design.

| Category | Examples | Notes |
|---|---|---|
| Process parameters | Injection and packing pressure profile (time-dependent), resin melt temperature, cooling time, mold temperature. | Primary control variables. Strongly influence defect formation, dimensional accuracy, and cycle time. |
| Material/geometry variables | Resin type and filler contents, part geometry (rib thickness, wall thickness, aspect ratio). | Affect flow behavior, cooling rate, mechanical properties. Geometry must satisfy moldability constraints. |
| Environmental (contextual) variables | Ambient temperature and humidity, machine conditions. | External factors shifting optimal process settings. Important for robustness and generalization. |
| Objectives | Part defects, dimensional accuracy, mechanical strength, cycle time, energy consumption per cycle. | Typical multi-objective optimization targets in industrial designs. |
| Manufacturing constraints | Machine operating limits (e.g., max pressure, clamping force), energy consumption and policy limits, resin thermal window, minimum part feature size, geometric moldability. | Define feasible design space. Must be considered during problem definition and optimization. |